\documentclass[10pt,twocolumn,letterpaper]{article}

\usepackage{iccv}
\usepackage{times}
\usepackage{epsfig}
\usepackage{graphicx}
\usepackage{amsmath}
\usepackage{amssymb}
\usepackage{xcolor}
\usepackage[normalem]{ulem}
\usepackage{comment}
\usepackage{booktabs}

\usepackage[breaklinks=true,bookmarks=false]{hyperref}

\iccvfinalcopy 

\def\iccvPaperID{5196} 

\newcommand\Mark[1]{\textsuperscript{#1}}

\ificcvfinal\fi

\begin{document}


\title{Structured Modeling of Joint Deep Feature and Prediction Refinement for Salient Object Detection}

\author{
    {Yingyue Xu\Mark{1}, Dan Xu\Mark{2}, Xiaopeng Hong\Mark{3,7,1}, Wanli Ouyang\Mark{4}, Rongrong Ji\Mark{5,7}, Min Xu\Mark{6}, Guoying Zhao\Mark{1}\thanks{Guoying Zhao is the corresponding author.}}\vspace{6pt}
    \\\Mark{1}University of Oulu\thanks{This work is supported by the Academy of Finland ICT 2023 project (313600), Tekes Fidipro program (Grant No.1849/31/2015), Business Finland project (Grant No.3116/31/2017), Infotech Oulu, and National Natural Science Foundation of China (Grant No.61772419). Computational resources are supported by CSC-IT Center for Science, Finland and Nvidia.} \quad 
    \Mark{2}University of Oxford \quad 
    \Mark{3}Xi'an Jiaotong University \quad
    \\\Mark{4}SenseTime Computer Vision Group, The University of Sydney \quad 
    \\\Mark{5}Xiamen University \quad 
    \Mark{6}University of Technology Sydney \quad 
    \Mark{7}Peng Cheng Laborotory \quad 
}

\maketitle
\thispagestyle{empty}

\begin{abstract}
Recent saliency models extensively explore to incorporate multi-scale contextual information from Convolutional Neural Networks (CNNs). Besides direct fusion strategies, many approaches introduce message-passing to enhance CNN features or predictions. However, the messages are mainly transmitted in two ways, by feature-to-feature passing, and by prediction-to-prediction passing. In this paper, we add message-passing between features and predictions and propose a deep unified CRF saliency model . We design a novel cascade CRFs architecture with CNN to jointly refine deep features and predictions at each scale and progressively compute a final refined saliency map. We formulate the CRF graphical model that involves message-passing of feature-feature, feature-prediction, and prediction-prediction, from the coarse scale to the finer scale, to update the features and the corresponding predictions. Also, we formulate the mean-field updates for joint end-to-end model training with CNN through back propagation. The proposed deep unified CRF saliency model is evaluated over six datasets and shows highly competitive performance among the state of the arts.

\end{abstract}

\section{Introduction}
Visual saliency, born from psychology~\cite{koch1987shifts}, refers to the attentional selection process on the scenes by the human visual system (HVS). At its early stage, saliency detection models focus on highlighting the most conspicuous regions or eye fixations on a scene~\cite{itti1998model,harel2007graph,judd2009learning,achanta2009frequency,xyy2018integration}. Later, the connotation of saliency is extended to object-level prediction by emphasizing the most outstanding objects. As a result, many salient object detection models are proposed~\cite{wei2012geodesic,yang2013saliency,mairon2014closer,jiang2015generic,liu2017hierarchical,wang2015deep,liu2016dhsnet,hou2017saliency,li2016visual,xyy2018saliency}, which may have a broad range of potential computer vision applications, such as segmentation~\cite{rahtu2010segmenting}, image cropping~\cite{santella2006gaze}, image fusion~\cite{han2013fast}, image classification~\cite{wang2010locality}, crowd counting~\cite{mawei2019bayloss}, video compression~\cite{guo2010novel}, \etc.

\begin{figure}[t]
\begin{center}
\begin{minipage}{0.32\linewidth}
     \centerline{\includegraphics[width=1in]{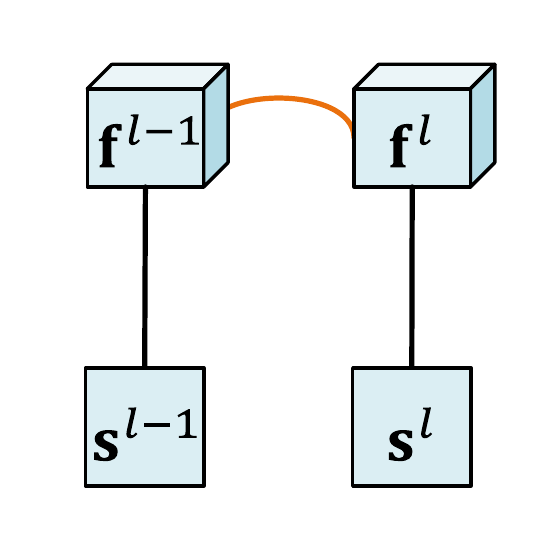}}
     \vspace{-5pt}
     \centerline{(a)}
\end{minipage}
\begin{minipage}{0.32\linewidth}
     \centerline{\includegraphics[width=1in]{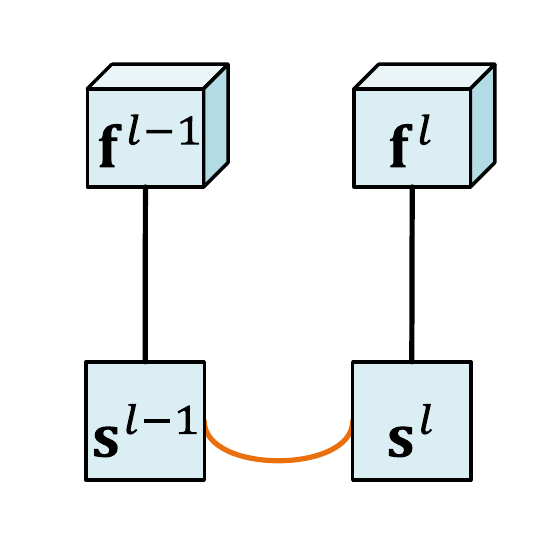}}
     \vspace{-5pt}
     \centerline{(b)}
\end{minipage}
\begin{minipage}{0.32\linewidth}
    \centerline{\includegraphics[width=1in]{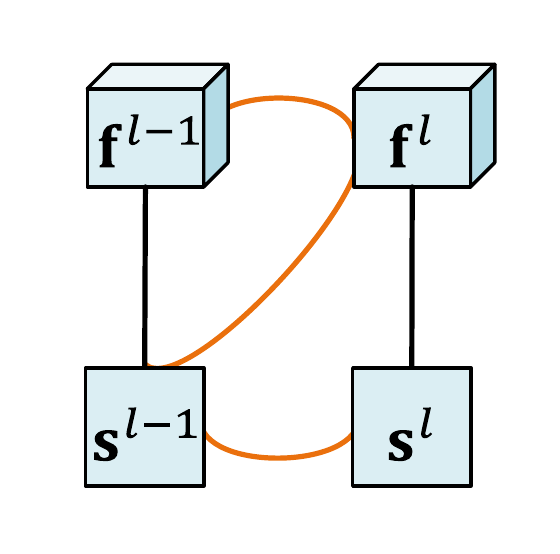}}
    \vspace{-5pt}
    \centerline{(c)}
\end{minipage}
\vspace{5pt}
\caption{Message-passing of (a) feature and feature, (b) prediction and prediction, and (c) \emph{joint} feature and prediction. 
$\textbf{f}^l$ and $\textbf{s}^l$ refer to the features and the corresponding prediction map at the $l$-th scale respectively. The orange curves indicate there are message-passing between two nodes.}
\vspace{-10pt}
\label{fig:passing}
\end{center}
\end{figure}

Recently, CNN based saliency models extensively explore to incorporate multi-scale contextual information. Besides directly fusing feature representations, recent approaches introduce message-passing to refine the multi-scale CNN contexts. Zhang~\etal~\cite{zhang2018bi} propose a gated bi-directional message-passing module to pass messages among \emph{features}, and thus to reinforce the multi-scale CNN features for saliency detection (Figure~\ref{fig:passing}-a). Other models pass messages among \emph{predictions} based on the conditional random field (CRF). For instance, recent saliency models~\cite{li2016deep,hou2017saliency,li2017instance} tend to adopt Dense-CRF~\cite{krahenbuhl2011efficient} that models highly structured message-passing between pixels on prediction maps, as a post-processing method for saliency refinement. Further, Xu~\etal~\cite{xu2017multi} introduce the multi-scale CRF to model message-passing between multi-scale prediction maps for depth estimation (Figure~\ref{fig:passing}-b). 

\begin{figure}[t]
\begin{center}
\scalebox{1}{\includegraphics[width=3.3in]{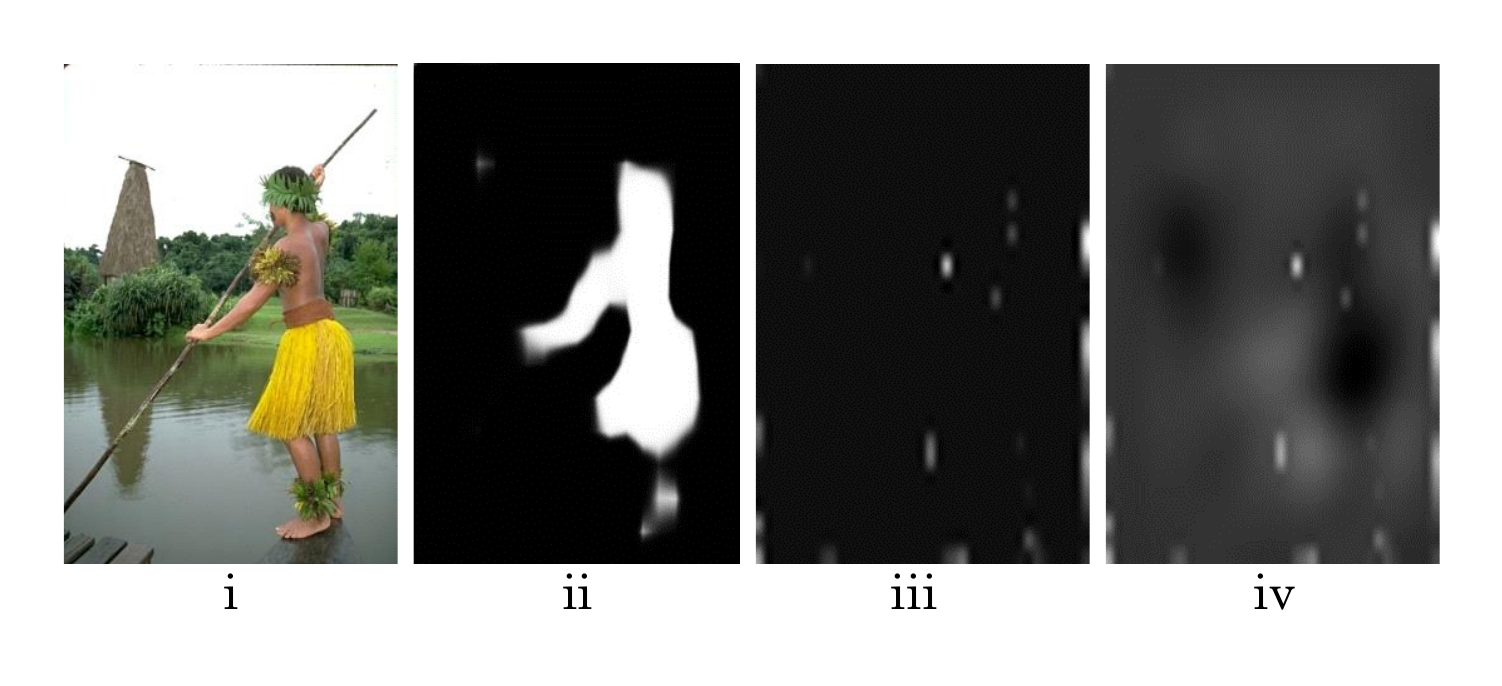}}
\vspace{-20pt}
\caption{From left to right columns are (i) input image; (ii) prediction map $\textbf{s}^2$; (iii) selected feature map from $\textbf{f}^3$; and (iv) the corresponding estimated feature map with message-passing between features and predictions. The backbone is detailed in Section~\ref{sec:implementation}.} 
\vspace{-15pt}
\label{fig:f-s-example}
\end{center}
\end{figure}

After close inspection, we notice that there are clear influences between features and predictions, and thus propose a joint feature and prediction refinement model by performing an extra message-passing between features and predictions. As in Figure~\ref{fig:passing}-c, messages are passing between features and features, predictions and predictions, and features and predictions. The motivations are two-fold:

Firstly, predictions may provide necessary contextual information to features. As the quality of the multi-channel feature maps may vary, the prediction map at the lower level can provide more spatial details to the features. As in Figure~\ref{fig:f-s-example}, the selected feature map in the third column shows inferior quality. Via message-passing between features and predictions, the inferior feature map is enhanced with spatial and shape information from the lower level prediction map. Then, the reinforced deep features will further improve the corresponding prediction map. Secondly, building message-passing between features and predictions facilitates efficient model training. During back propagation, the loss error may be transmitted slowly to impact the lower level features. Modeling message-passing between features and predictions is crucial to strengthen the connection of the model for efficient model training.  

\begin{figure*}[t]
\begin{center}
\scalebox{1}{\includegraphics[width=6.9in]{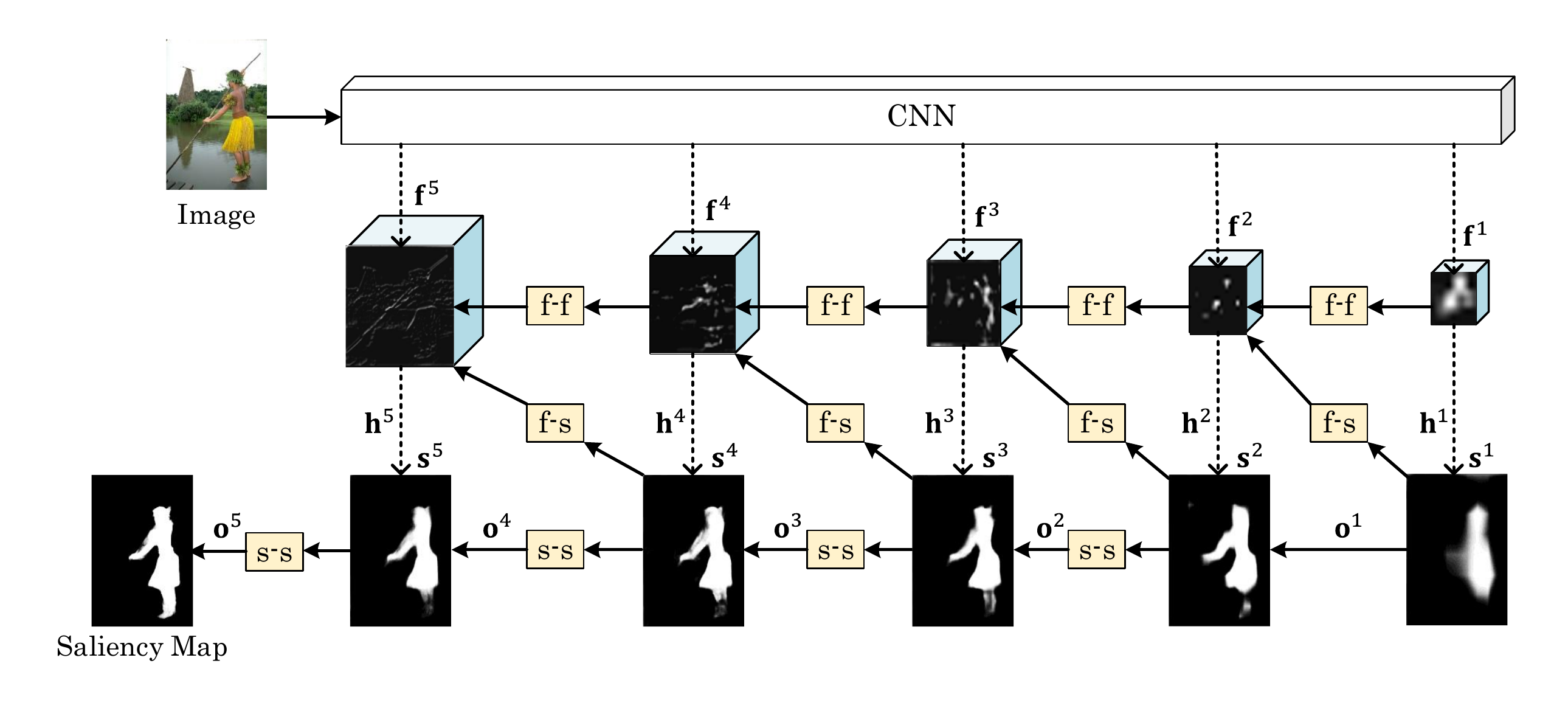}}
\vspace{-20pt}
\caption{Framework of the proposed deep unified CRF saliency model for jointly modeling structural deep features and predictions. Multi-scale features ($\textbf{f}^1$$\cdots$$\textbf{f}^5$) and the corresponding prediction maps ($\textbf{s}^1$$\cdots$$\textbf{s}^5$) are extracted from the backbone CNN. At each scale, a CRF block is embedded to jointly refine features and prediction maps with message-passing between features and features ({f}-{f}), features and predictions ({f}-{s}), and predictions and predictions ({s}-{s}). ``$\textbf{h}^1$$\cdots$$\textbf{h}^5$'' and ``$\textbf{o}^1$$\cdots$$\textbf{o}^5$'' are the estimated features and predictions at each scale respectively.  ``$\textbf{f}^1$$\cdots$$\textbf{f}^5$'' correspond to ``pool5a'', ``conv5{\_}3'', ``conv4{\_}3'', ``conv3{\_}3'' and ``conv2{\_}2'' in the enhanced HED~\cite{xie2015holistically} structure, while ``$\textbf{s}^1$$\cdots$$\textbf{s}^5$'' are ``upscore-dsn6'', ``upscore-dsn5'', ``upscore-dsn4'', ``upscore-dsn3'', and ``upscore-dsn2''. The dashed arrows omit the details within the backbone CNN. Figure~\ref{fig:CRF_detail} details the implementation within each CRF block.} 
\vspace{-15pt}
\label{fig:frame}
\end{center}
\end{figure*}

Thus, in this paper, we propose a deep unified CRF saliency model that formulates messages passing with CRFs for joint feature and prediction refinement. We design a novel deep cascade CRFs architecture that is seamlessly incorporated with CNN to integrate and refine multi-scale deep features and predictions for a refined saliency map. At each scale, a CRF block is embedded that takes the features and predictions from the lower scale as observed variables to estimate the hidden features and predictions at the current scale. Within each CRF inference, feature-feature, feature-prediction and prediction-prediction messages passing are built. Then, the output refined features and the prediction map are incorporated into the CRF block at the next scale. Thus, a series of CRFs are constructed in a cascade flow and progressively learn a unified saliency map from the coarse scale to the finer scale. Moreover, the CRF inference is formulated with mean-field updates that allow jointly end-to-end training with CNN by back-propagation. The framework of the model is presented in Figure~\ref{fig:frame}. 



The contributions of this paper are two-fold:

\begin{itemize}
\itemsep0em 
\vspace{-5pt}
  \item We propose a cascade CRFs architecture that is seamlessly incorporated with the backbone CNN to progressively integrate and refine multi-scale contexts from CNNs for salient object detection.
  \item We model structural information of deep features and deep predictions into a unified CRF model for joint refinement and develop the mean-field approximation inference that supports end-to-end model training through back propagation.
\vspace{-3pt}
\end{itemize}

\section{Related Works}\label{sec:relatework}
\par\noindent\textbf{Saliency Models Based on Multi-scale CNNs} 
In the past few years, a broad range of saliency models based on CNNs has been proposed. Early deep saliency models benefit from adjusting the inputs to VGG~\cite{simonyan2014very}, of which the inputs are either multi-scale resized images~\cite{liu2015predicting} or global and local image segments~\cite{li2016visual,wang2015deep,zhao2015saliency,kim2016shape}. Recently, deep saliency models extensively take the advantages of the multi-scale contexts from CNNs and adopt variable fusion strategies to produce the saliency map. A hierarchical architecture can effectively refine the CNN side outputs from coarse to fine scales~\cite{liu2016dhsnet,zhang2017learning,liu2018picanet}. PiCANet~\cite{liu2018picanet} hierarchically embeds global and local contexts. Moreover, some saliency models adopt recurrent or cascade structures to progressively learn saliency maps from coarse to fine scales~\cite{wang2016saliency,kuen2016recurrent,zhang2018progressive,chen2018reverse}. Zhang~\etal~\cite{zhang2018progressive} introduce a multi-path recurrent feedback scheme to progressively enhance the saliency prediction map. RA~\cite{chen2018reverse} introduces reverse attention with side-output residual learning to refine the saliency map in a top-down manner. Also, skip connections are widely applied to integrate prediction maps from CNNs~\cite{zhang2017amulet,hou2017saliency}. DSS~\cite{hou2017saliency} adopts short connections to the side output layers of CNNs to fuse multiple prediction maps. In this work, we aim at integrating multi-scale deep features and deep predictions to boost performance. 


\par\noindent\textbf{Fully Connected CRFs} As the conditional random field (CRF) is a flexible graphical model in incorporating label agreement assumptions into inference functions, it has been widely adopted for labeling refinement tasks. Several deep saliency models~\cite{li2016deep,hou2017saliency,li2017instance} take the advantages of CRF inference and apply a fully connected Dense-CRF~\cite{krahenbuhl2011efficient} to CNN as a post-processing method for refinement. Dense-CRF works on the \emph{discrete} semantic segmentation, which yields an effective iterative message-passing algorithm using mean-field theory. The mean-field approximation can be performed using highly efficient Gaussian filtering in feature space, reducing the complexity from quadratic to linear. However, Dense-CRF parameters are pre-selected by cross validations from a large number of trials and thus is disconnected from the training of CNNs. Zheng~\etal~\cite{zheng2015conditional} firstly formulate the CRF inference on top of CNN predictions that enables joint model training via back propagation for semantic segmentation. To solve depth estimation in the \emph{continuous} domain, Xu~\etal~\cite{xu2017multi, xu2017monocular} introduce the continuous CRF that incorporates multi-scale CNN prediction maps. Later, Xu~\etal~\cite{xu2017learning} also propose the attention gated CRF that allows message-passing among the \emph{continuous} features for contour prediction. Chu~\etal~\cite{chu2016crf} pass messages among features for pose estimation.

All these models formulate CRF with message-passing only among features or among predictions, while we first formulate the \emph{continuous} feature variables and the \emph{discrete} prediction variables into a deep unified CRF model. The new CRF formulation provides explainable solutions for the features, the predictions and the interactions among them, leading to distinct model formulation, inference, and neural network implementation.

\section{The Deep Unified CRF Saliency Model}\label{sec:model}

\par\noindent\textbf{Formulation.} Given an input image $\textbf{I}$ of $N$ pixels, suppose that a backbone CNN network computes $L$ scales of deep feature maps $\textbf{F}=\{\textbf{{f}}^l\}_{l=1}^L$, where $\textbf{{f}}^l=\{{\text{f}}_{i,m}^{l}\}_{i=1,m=1}^{N,M}$ consists a set of $M$ feature vectors. Accordingly, $L$ scales of prediction maps $\textbf{S}=\{\textbf{{s}}^l\}_{l=1}^L$ can be computed, where $\textbf{{s}}^l=\{{\text{s}}_i^l\}_{i=1}^N$. The ground truth saliency map corresponding to the input image is denoted as $\textbf{{g}}=\{\text{g}_i\}_{i=1}^N$, and each element $\text{g}_i$ takes binary values of 1 or 0. 

We formulate the CRF inference to jointly refine multi-scale features and predictions. The objective is to approximate the hidden multi-scale deep feature maps $\textbf{H}=\{\textbf{{h}}^l\}_{l=1}^L$ and the hidden multi-scale prediction maps $\textbf{O}=\{\textbf{{o}}^l\}_{l=1}^L$. In particular, at the $l$-th scale, the observed variables are the features $\textbf{{f}}^{l-1}$, $\textbf{{f}}^{l}$ and the prediction $\textbf{{s}}^{l-1}$, and the objective is to estimate the corresponding $\textbf{{h}}^{l}$ and $\textbf{{o}}^{l}$. With a cascade flow of a series of CRFs, the side outputs are progressively refined from coarse ($l=1$) to fine ($l=L$). The refined prediction map $\textbf{{o}}^L$ is the final saliency map.

The conditional distribution of the CRF at the $l$-th scale is defined as follow:
\begin{equation} \label{eq:crf}
P(\textbf{{h}}^l,\textbf{{o}}^l|\textbf{I},\Theta)=\frac{1}{Z(\textbf{I},\Theta)} \exp\Big\{-E\left(\textbf{{h}}^l,\textbf{{o}}^l,\textbf{I},\Theta\right)\Big\},
\end{equation}

\noindent where $\Theta$ refers to the relative parameters. The energy function $E = E(\textbf{{h}}^l,\textbf{{o}}^l,\textbf{I},\Theta)$ is formulated as follow:

\begin{equation} \label{eq:energy}
\setlength\abovedisplayskip{0pt} 
\begin{aligned}
E& =\sum_i\phi_h(\text{h}_i^l,\text{f}_i^l)+\sum_i\phi_o(\text{s}_i^l,\text{o}_i^l)+
\sum_{i{\neq}j}\psi_h(\text{h}_i^l,\text{h}_j^{l-1}) +\\&\sum_i\psi_{hs}(\text{h}_i^l,\text{o}_i^{l-1})+\sum_{i\neq j}\psi_o(\text{o}_i^l,\text{o}_j^l).
\end{aligned}
\setlength\belowdisplayskip{0pt}
\end{equation}
The first term of Eq.~\ref{eq:energy} is a feature level unary term corresponding to an isotropic Gaussian: $\phi_h(\text{h}_i^l,\text{f}_i^l) = -\frac{\alpha_i^l}{2}||\text{h}_i^l-\text{f}_i^l||^2$ where ${\alpha_i^l}>0$ is a weighting factor. 
The second term is a prediction level unary term, $\phi_o(\text{s}_i^l,\text{o}_i^l) = ||\text{s}_i^l-\text{o}_i^l||^2$. The third term is a feature level pairwise term describing the potential between features, where
\begin{equation}
\psi_h(\text{h}_i^l,\text{h}_j^{l-1})= \text{h}_i^l\text{W}_{i,j}^{l,l-1}\text{h}_j^{l-1},
\end{equation}
where $\text{W}_{i,j}^{l,l-1}\in \mathbb{R}^{M\times M}$ is a bilinear kernel. 
The fourth term is a feature level pairwise term defining the potential between features and predictions, where
\begin{equation}
\psi_{hs}(\text{h}_i^l,\text{o}_i^{l-1})=\text{h}_i^l\text{V}_{i,j}^{l,l-1}\text{o}{'}_j^{l-1}, 
\end{equation}
where $\text{V}_{i,j}^{l,l-1}\in \mathbb{R}^{M\times M}$ is also a bilinear kernel to couple the features and the predictions. $\text{o}'^{l-1}$ denotes a concatenation of $M$ prediction maps $\text{o}^{l-1}$. The fifth term is a prediction level pairwise term defining the potential between the predictions as follows:
\begin{equation}
\psi_o(\text{o}_i^l,\text{o}_j^l)=\beta_1 K_{i,j}^{1} ||\text{o}_i^l-\text{o}_j^l||^2+\beta_2 K_{i,j}^{2} ||\text{o}_i^l-\text{o}_j^l||^2.
\end{equation}
$K_{i,j}^{1}$ and $K_{i,j}^{2}$ are Gaussian kernels that measure the relationship between two pixels. Specifically, $K_{ij}^{1}$ is the similarity kernel measuring the appearance similarity between two pixels as $K_{ij}^{1} = \nu_1\exp(-\frac{\|p_i-p_j\|^2}{2\sigma_\alpha^2}-\frac{\|I_i-I_j\|^2}{2\sigma_\beta^2})$ and $K_{ij}^{2}$ is the proximity kernel that measures the spatial relationship between two pixels as $K_{ij}^{2} = \nu_2\exp(-\frac{\|p_i-p_j\|^2}{2\sigma_\gamma^2})$. $\nu_1$ and $\nu_2$ are the contributions of each Gaussian kernel, respectively.




\par
\begin{figure}[t]
\begin{center}
\scalebox{1}{\includegraphics[width=3.2in]{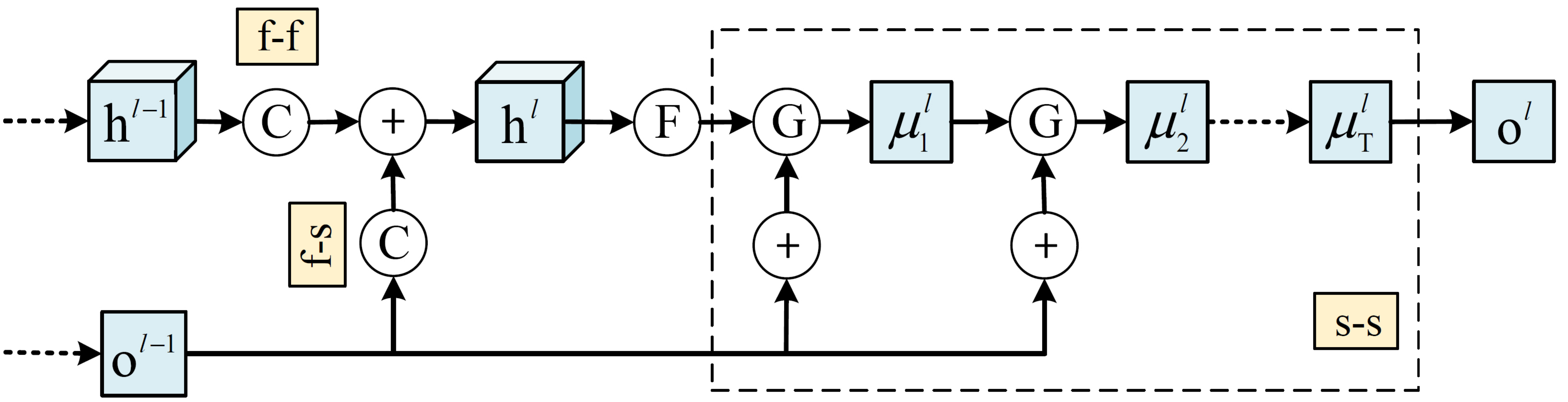}}
\caption{Details of the mean-field updates within CRF. The circled symbols indicate message-passing operations within the CRF block. (i) Message-passing to estimate $\textbf{h}^l$ by convolutions (Eq.~\ref{eq:mean}): ``C'' indicates a convolutional layer followed by the corresponding deconvolutional layer, crop layer and a scale layer. (ii) Message-passing to estimate $\textbf{o}^l$ with Gaussian pairwise kernels in $T$ iterations (Eq.~\ref{eq:output}): ``G'' means the Gaussian filtering. ``F'' is the process of computing a prediction map by a convolutional layer with kernel size $1\time 1$ followed with the corresponding deconvolutional layer and a crop layer. ``+'' refers to element-wise sum.} 
\vspace{-8pt}
\label{fig:CRF_detail}
\end{center}
\end{figure}

\noindent\textbf{Inference.}~We perform the mean-field approximation to estimate a distribution $q(\textbf{{h}}^l,\textbf{{o}}^l|\textbf{I},\Theta)=\prod_{i=1}^{N}q_{i,l}(\textbf{I},\Theta|\text{h}_i^l,\text{o}_i^l)$ that is an approximation to $P(\textbf{{h}}^l,\textbf{{o}}^l|\textbf{I},\Theta)$ by minimizing the Kullback-Leiber divergence~\cite{ristovski2013continuous}. By considering $J_{i,l}=\log q_{i,l}(\textbf{{h}}^l,\textbf{{o}}^l|\textbf{I},\Theta)$ and rearranging its expression into an exponential form, the mean-field updates can be derived as:
\begin{align} \label{eq:mean}
\begin{split} 
\bar{\text{h}}_i^l=&\frac{1}{\alpha_i^l}\Big(\alpha_i^l\text{f}_i^l+\sum_{l{\neq}l-1}\sum_{i{\neq}j}\text{W}_{i,j}^{l,l-1}\text{h}_j^{l-1}\\
&+\sum_{l{\neq}l-1}\sum_{i{\neq}j}\text{V}_{i,j}^{l,l-1}\text{o}{'}_j^{l-1}\Big).
\end{split}
\end{align}

\begin{equation} \label{eq:norm}
\rho_i^l=1+2\Big(\beta_1\sum_{j\neq i}K_{ij}^{1}+\beta_2\sum_{j\neq i}K_{ij}^{2}\Big),
\end{equation}
\begin{equation} \label{eq:output}
\mu_i^l=\frac{o_i^l}{\rho_i^l}+\frac{2}{\rho_i^l}\Big(\beta_1\sum_{j\neq i}K_{ij}^{1}\mu_j^l+\beta_2\sum_{j\neq i}K_{ij}^{2}\mu_j^l\Big).
\end{equation}
Where Eq.~\ref{eq:mean} and~\ref{eq:output} represent the mean-field inference of the estimated latent feature and prediction variables, respectively. Eq.~\ref{eq:norm} is the variance of the mean-field approximated distribution used as the normalization factor in Eq.~\ref{eq:output}. At the $l$-th scale, the optimal ${\textbf{o}}^l$ can be approximated by mean-field updates of $T$ iterations on the prediction level. At each time $t$ of the mean-field iteration, an estimated saliency map $\mu_t^{l}$ can be approximated. After $T$ mean-field iterations, the estimated prediction map $\mu_T^{l}$ is regraded as the estimation of $\textbf{o}^l$ from the CRF at the $l$-th scale. The details of the mean-field updates are presented in Figure~\ref{fig:CRF_detail}.

In the cascade flow, the observation $\textbf{s}^l$ is obtained via integrating the prediction map $\textbf{s}^l$ and the estimated map $\textbf{o}^{l-1}$ from the CRF at the previous scale, \ie, $\text{s}_i^l=\text{s}_i^l+\text{o}_i^{l-1}$. 

\par\noindent\textbf{Mean-field Iteration with Neural Networks.}
The inference of the CRF block is based on the mean-field approximation, which can be implemented as a stack of CNN layers to facilitate jointly training as in Figure~\ref{fig:CRF_detail}. The mean-field updates for Eq.~\ref{eq:mean} is implemented with convolutions. The similarity kernel $K^{\text{1}}$ and the proximity kernel $K^{\text{2}}$ in Eq.~\ref{eq:norm} and~\ref{eq:output} are computed based on permutohedral lattice~\cite{adams2010fast} to reduce the computational cost from quadratic to linear~\cite{ristovski2013continuous}. The weighting of $\beta_1$ and $\beta_2$ is convolved with an 1$\times$1 kernel. By combining the outputs, the normalization matrix $\rho^l$ and the corresponding $\mu^l$ can be computed. The weights $\beta_1$ and $\beta_2$ are obtained by back propagation. 


\section{Implementation} \label{sec:implementation}

\par\noindent\textbf{Training Data.}
As many state-of-the-art saliency models use the MSRA-B dataset~\cite{liu2011learning} as the training data~\cite{hou2017saliency,li2016deep,wang2016saliency,liu2016dhsnet}, we also follow the same training protocol as in ~\cite{hou2017saliency,li2016deep} to optimize the deep unified CRF model, for fair comparisons. The MSRA-B dataset consists of 2,500 training images, 500 validation images, and 2000 testing images. The images are resized to 240$\times$320 as the input to the data layer. Horizontal flipping is used for data augmentation such that the number of training samples is twice as large as the original number.

\par\noindent\textbf{Baseline Model.} In order to learn high quality feature maps and to fairly compare the proposed deep unified CRF model with the state of the arts, the front-end CNN is based on the implementation of DSS~\cite{hou2017saliency} with the enhanced HED~\cite{xie2015holistically} structure. Latest state-of-the-art saliency models, \eg, DSS~\cite{hou2017saliency} and RA~\cite{chen2018reverse} both adopt such front-end network to extract multi-scale side outputs. The only difference is that we discard the side output prediction maps computed from the layer ``conv1{\_}2'', which is used as the sixth side output map by DSS~\cite{hou2017saliency} and RA~\cite{chen2018reverse}. Thus, totally five scales of side outputs are extracted. Specifically, the multi-scale features ``$\textbf{f}^1$$\cdots$$\textbf{f}^5$'' correspond to ``pool5a'', ``conv5{\_}3'', ``conv4{\_}3'', ``conv3{\_}3'' and ``conv2{\_}2'' in the enhanced HED~\cite{xie2015holistically} structure, while the corresponding prediction maps ``$\textbf{s}^1$$\cdots$$\textbf{s}^5$'' are ``upscore-dsn6'', ``upscore-dsn5'', ``upscore-dsn4'', ``upscore-dsn3'', and ``upscore-dsn2'' respectively.
\par\noindent\textbf{Optimization.} To reduce training time, the proposed deep unified CRF model is optimized with two stages, including a pre-training and an overall optimization. 

In the pre-training stage, we firstly optimize the model by adding feature-feature and feature-prediction messages passing to the front-end CNN. The parameters $\Delta$ of the networks and the scale-specific parameters ${\varepsilon}$=$\{\epsilon_l\}_{l=1}^L$ are trained by minimizing the standard sigmoid cross-entropy loss. The output score maps at each scale are optimized respectively. Thus, the loss function is as follow:
\begin{equation*} 
\setlength\abovedisplayskip{5pt} 
\mathcal{L}_{\text{Stage1}}^l(\Delta, \epsilon_l)=-\sum_{i=1}^\mathcal{N}\Big(\text{g}_i\log(\text{s}_i^l)+(1-\text{g}_i)\log(1-\text{s}_i^l)\Big),
\setlength\belowdisplayskip{5pt} 
\end{equation*}
\noindent where $\text{g}_i$ is the $i$-th ground-truth label and $\text{s}_i^l$ is the $i$-th pixel on the prediction map at the $l$-th scale. $\mathcal{N}$ refers to the total number of image pixels over the training set. 

In the second stage, overall parameter optimization is performed to the whole network. Still, the parameters \{$\Delta, \varepsilon$\} trained from the pre-training stage will be jointly optimized with the parameters $\beta=\{\beta_1^l,\beta_2^l\}_{l=1}^L$ for prediction-prediction message-passing at each scale. The sigmoid cross-entropy loss function is computed for the final scale $L$ as follow: 
\begin{equation*} 
\setlength\abovedisplayskip{5pt} 
\mathcal{L}_{\text{Stage2}}(\Delta,\varepsilon,\beta)=-\sum_{i=1}^\mathcal{N}\Big(\text{g}_i\log({{\text{o}}}_i^L)+(1-\text{g}_i)\log(1-{\text{{o}}}_i^L)\Big).
\setlength\belowdisplayskip{5pt} 
\end{equation*}

\par\noindent\textbf{Parameters.} In this work, the VGG-16~\cite{simonyan2014very} is adopted to initialize the parameters for the pre-training stage, and the front-end CNN is finetuned.  The parameters for the pre-training stage are set as: batch\_size (1), learning rate (1e-9), max\_iter (14000), weight decay (0.0005), momentum (0.9), iter\_size (10). The learning rate is decreased by 10\% when the training loss reaches a flat.

In the second training stage, the parameters learned from the pre-training stage are optimized with a learning rate of 1e-12, while the parameters for prediction-prediction message-passing are learned with the learning rate as 1e-8. Another 10 epochs are trained for the overall optimization.

All the implementation is based on the public Caffe library~\cite{jia2014caffe}. The Gaussian pairwise kernels are implemented based on continuous CRF~\cite{xu2017multi}. The GPU for training acceleration is the Nvidia Tesla P100. The pre-training takes about 6 hours and the overall training takes about 14 hours.

\begin{figure}
\begin{center}
\scalebox{0.5}{\includegraphics[width=6.7in]{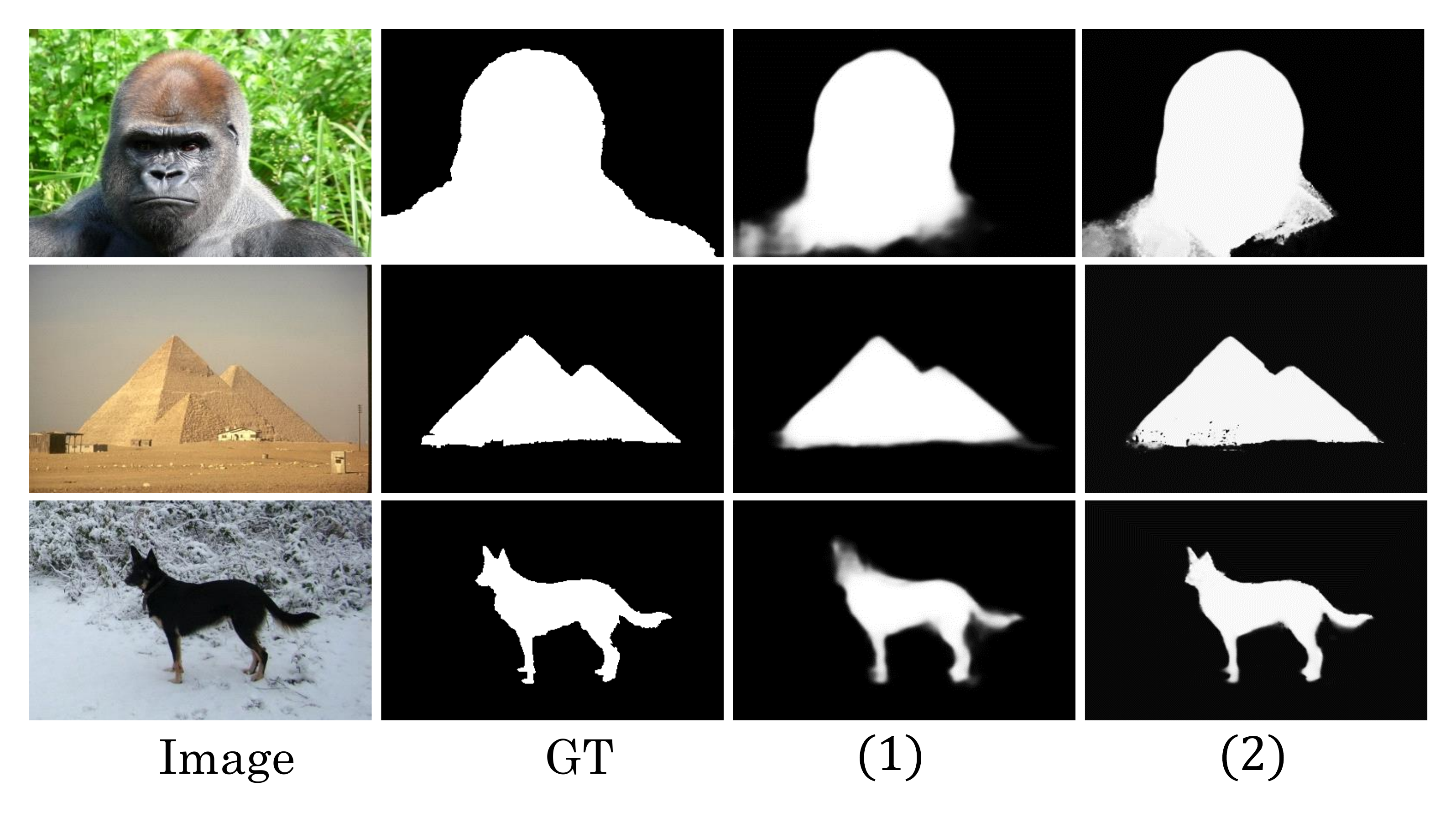}}
\caption{Prediction maps $\textbf{o}^5$ from the deep unified CRF model with only message-passing bewteen predictions with emphasis on (1) proximity ($T=6$, $\sigma_\alpha=1$, $\sigma_\beta=10$, $\sigma_\gamma=10$) and (2) similarity ($T=6$, $\sigma_\alpha=10$, $\sigma_\beta=10$, $\sigma_\gamma=1$). Example images are selected from the ECSSD dataset with the ground truth.}
\label{fig:param}
\vspace{-6pt}
\end{center}
\end{figure}

\section{Experiments} \label{sec:experiment}

\subsection{Datasets}\label{subsec:dataset}
For comprehensive comparisons, the proposed deep unified CRF saliency model is evaluated over six datasets, including: MSRA-B~\cite{liu2011learning}, ECSSD~\cite{yan2013hierarchical}, PASCAL-S~\cite{li2014secrets}, DUT-OMRON~\cite{yang2013saliency}, HKU-IS~\cite{li2016visual} and iCoseg~\cite{batra2010icoseg}. MSRA-B is the training dataset consisting of 5000 images. ECSSD contains a pool of 1000 images with even more complex salient objects on the scenes. PASCAL-S is a dataset for salient object detection consisting of a set of 850 images from PASCAL VOC 2010~\cite{pascal-voc-2010} with multiple salient objects on the scenes. DUT-OMRON dataset contains a large number of 5168 more difficult and challenging images. HUK-IS consists of 4447 challenging images and pixel-wise saliency annotation. ICoseg contains 643 images and each image may consist of multiple salient objects.

\subsection{Evaluation Metrics}
We employ two types of evaluation metrics to evaluate the performance of the saliency maps: F-measure and mean absolute error (MAE). When a given saliency map is slidingly thresholded from 0 to 255, a precision-recall (PR) curve can be computed based on the ground truth. F-measure is computed to count for the saliency maps with both high precision and recall:
\begin{equation} \label{eq:fmeasure}
F = \frac{\left(1+{b}^2\right)\cdot \texttt{precision}\cdot \texttt{recall}}{b^2 \cdot \texttt{precision} + \texttt{recall}},
\end{equation}
where $b^2$ is set as 0.3~\cite{achanta2009frequency} to emphasize the precision. In this paper, the Max F-measure is evaluated. MAE~\cite{perazzi2012saliency} measures the overall pixel-wise difference between the saliency map $\texttt{sal}$ and the ground truth $\texttt{gt}$: 
\begin{equation} \label{eq:mae}
\texttt{MAE} = \frac{1}{N}\sum_{i=1}^{N} {\left|\texttt{sal}(i)-\texttt{gt}(i)\right|}.
\end{equation}


\begin{table}
\begin{center}
    \begin{tabular}{c|ccccc|cc}
    \toprule
    $T$	&$\sigma_\alpha$&$\sigma_\beta$	&$\sigma_\gamma$	&$\nu_1$	&$\nu_2$	&F-measure&MAE\\\midrule\midrule
    3	&1	&1	&1	&1	&1	&0.892	&0.071\\
    6	&1	&1	&1	&1	&1	&0.892	&0.071\\
    6	&10	&10	&10	&1	&1	&\underline{0.909}	&0.071\\
    6	&10	&1	&1	&1	&1	&0.896	&0.084\\
    6	&1	&10	&1	&1	&1	&0.893	&\underline{0.070}\\
    6	&1	&1	&10	&1	&1	&0.892	&\underline{0.070}\\
    6	&10	&10	&1	&1	&1	&\textbf{0.910}	&0.094\\
    6	&1	&10	&10	&1	&1	&0.894	&\textbf{0.069}\\
    6	&10	&1	&10	&1	&1	&0.896	&0.095\\
    6	&1	&1	&1	&3	&5	&0.892	&0.071\\
    6	&1	&1	&1	&5	&3	&0.892	&0.071\\
    6	&1	&1	&1	&1	&1	&0.892	&0.071\\
    \bottomrule
    \end{tabular}
    \vspace{5pt}
    \caption{One CRF block with only message-passing between predictions at scale 5 is jointly trained with the backbone CNN for 10 epochs. The model is tested on ECSSD dataset. $T$ refers to the number of mean-field iterations. The similarity kernel is controlled by $\sigma_\alpha$ and $\sigma_\beta$ and the weight $\nu_1$, while the proximity kernel is controlled by $\sigma_\gamma$ and the weight $\nu_2$. The best performances are in bold while the second best results are underlined.}
    \vspace{-15pt}
    \label{tbl:param}
\end{center}
\end{table} 

\subsection{Model Analysis}

\par\noindent\textbf{Gaussian Pairwise Kernels.} 
Recall that $\nu_1$, $\sigma_\alpha$ and $\sigma_\beta$ are pre-defined parameters to control the bandwidth of $K^{1}$, while $\nu_2$ and $\sigma_\gamma$ control the bandwidth of $K^{2}$ for prediction level message-passing. Various schemes of parameter settings are experimented as in Table~\ref{tbl:param}, of which one CRF block at scale 5 is jointly trained with the front-end CNN for 10 epochs. It can be perceived that when the similarity $K^{1}$ is emphasized, the output map receives better F-measure; when the proximity $K^{2}$ is emphasized, the MAE sharply reduces. Meanwhile, different settings of $\nu_1$ and $\nu_2$ result in the same performance. This is because that the mean-field iterations learn the weights of the two Gaussian pairwise kernels $\beta_1$ and $\beta_2$ respectively, such that we can initialize $\nu_1$ and $\nu_2$ as 1\footnote{We set $\nu_1$ and $\nu_2$ as 1 in the following experimental descriptions.}. Thus, compared to Dense-CRF~\cite{krahenbuhl2011efficient}, our proposed CRF releases two pre-defined parameters.

\begin{table}[t]
\small
\begin{center}
    \begin{tabular}{c|ccccc}
    \toprule
    Scale ($l$)	&1	&2	&3	&4	&5\\\midrule
    Predictions ($\textbf{s}^l$)	&0.824	&0.864	&0.882	&0.883	&0.884\\
    Estimations (${\textbf{o}}^l$)	&	-&0.894	&0.915	&0.921	&0.921\\
    \bottomrule
    \end{tabular}
    \vspace{5pt}
    \caption{F-measure of prediction maps $\textbf{s}^l$ from each scale of the pre-trained backbone CNN and the estimated prediction maps $\textbf{o}^l$ at each scale of the deep unified CRF model, on ECSSD dataset.}
    \vspace{-15pt}
    \label{tbl:scale}
\end{center}
\end{table} 

\begin{table*}[thbp!]
\begin{center}
\small
    \begin{tabular}{c|c|ccccccccccc|c}
    \toprule
    {Dataset}	& {Metric}	& {DRFI}	& {MDF}	& {RFCN}	& {DHS}	& {Amulet}	& {UCF}	& {DCL$^+$}	& {MSR$^+$}	& {DSS}	& {DSS$^+$}	& {RA}	& {Ours}\\\midrule
    {MSRA-B}	&{maxF}	&0.851	&0.885	&-	&-	&-	&-	&0.916	&0.930	&0.920	&{0.928}	&\underline{0.931}	&\textbf{0.935}\\
    
    	&{MAE}	&0.123	&0.066	&-	&-	&-	&-	&0.047	&0.042	&0.043	&\underline{0.035}	&0.036	&\textbf{0.029}\\\midrule
    	
    {PASCAL-S}	&{maxF}	&0.690	&0.759	&0.829	&0.824	&0.832	&0.818	&0.822	&\underline{0.852}	&0.826	&0.831	&0.829	&\textbf{0.858}\\
    
    	&{MAE}	&0.210	&0.142	&0.118	&0.094	&0.100	&0.116	&0.108	&\textbf{0.081}	&0.102	&0.093	&0.101	&\underline{0.089}\\\midrule
    	
    {DUT-}	&{maxF}	&0.664	&0.694	&0.747	&-	&0.743	&0.730	&0.757	&0.785	&0.764	&0.781	&\underline{0.786}	&\textbf{0.802}\\
    
    {OMRON}	&{MAE}	&0.150	&0.092	&0.095	&-	&0.098	&0.120	&0.080	&0.069	&0.072	&0.063	&\underline{0.062}	&\textbf{0.057}\\\midrule
    
    {HKU-IS}	&{maxF}	&0.775	&0.860	&0.894	&0.892	&0.897	&0.888	&0.904	&\underline{0.916}	&0.900	&\underline{0.916}	&0.913	&\textbf{0.920}\\
    
    	&{MAE}	&0.146	&0.129	&0.088	&0.052	&0.051	&0.061	&0.049	&\textbf{0.039}	&0.050	&\underline{0.040}	&0.045	&\textbf{0.039}\\\midrule
    	
    {ECSSD}	&{maxF}	&0.784	&0.847	&0.899	&0.907	&0.914	&0.902	&0.901	&0.913	&0.908	&\underline{0.921}	&\underline{0.921}	&\textbf{0.928}\\
    
    	&{MAE}	&0.172	&0.106	&0.091	&0.059	&0.061	&0.071	&0.068	&0.054	&0.062	&\underline{0.052}	&0.056	&\textbf{0.049}\\\midrule
    	
    {ICoseg}	&{maxF}	&0.812	&0.838	&0.846	&0.851	&\textbf{0.899}	&0.884	&0.875	&0.871	&0.860	&0.872	&0.868	&\underline{0.890}\\
    
    	&{MAE}	&0.145	&0.101	&0.097	&0.070	&0.070	&0.068	&\underline{0.066}	&0.147	&0.075	&0.068	&0.082	&\textbf{0.062}\\
    \bottomrule
    \end{tabular}
    \vspace{5pt}
    \caption{Evaluation results on six dataset and with models DRFI~\cite{jiang2013salient}, MDF~\cite{li2015visual}, RFCN~\cite{wang2016saliency}, DHS~\cite{liu2016dhsnet}, Amulet~\cite{zhang2017amulet}, UCF~\cite{zhang2017learning}, DCL~\cite{li2016deep}, MSR~\cite{li2017instance}, DSS~\cite{hou2017saliency}, RA~\cite{chen2018reverse} and the deep unified CRF model. ``$+$'' marks the models utilizing Dense-CRF~\cite{krahenbuhl2011efficient} for post-processing. ``-'' means that the corresponding dataset is used as the training data. The evaluation on MSRA-B is performed on the testing set.}
    \label{tbl:f_m}
\end{center}
\end{table*} 

Figure~\ref{fig:param} presents three examples when the similarity $K^{{1}}$ or the proximity $K^{{2}}$ are emphasized respectively. When the proximity counts more, the output saliency maps are smoother. However, as the \textit{dog} example shows, the saliency objects possess more ambiguous object boundaries. The similarity kernel, however, emphasizes more on the image feature similarity, such that it is more sensitive to boundary division. But as the similarity may be too sensitive to the details on the image, this also results in some defects shown in the \textit{pyramids} example where the left corner of the pyramids contains many flaws. For evaluation, F-measure is based on thresholded segmentation to evaluate region similarity~\cite{fan2018salient}, while MAE calculates for pixel level accuracy. Thus, the emphasis on similarity gets better F-measure, while the emphasis on proximity gets better MAE.

\par\noindent\textbf{Scale-specific Gaussian Kernels.} We evaluate the F-measure of the estimated prediction maps ${\textbf{o}}^l$ at each scale of the deep unified CRF model with only message-passing between predictions. With Gaussian kernels $\sigma_\alpha=10$, $\sigma_\beta=10$, $\sigma_\gamma=1$, it results in the highest F-measure in Table~\ref{tbl:param}. According to Table~\ref{tbl:scale}, the quality of the estimated prediction maps ${\textbf{o}}^l$ from the CRFs continuously improves remarkably at scale 2, 3 and 4. However, the prediction maps $\textbf{s}^4$ and $\textbf{s}^5$ from the front-end CNN possess similar F-measure and the estimated map ${\textbf{o}}^5$ from the CRFs has almost no improvements. Thus, effective F-measure enhancement is performed only at scales from 2 to 4. Hence, it is sufficient for the cascade CRFs structure to integrate five side outputs from CNN rather than integrating six scales of maps by DSS~\cite{hou2017saliency} and RA~\cite{chen2018reverse} models.

In practice, the parameters of the first three CRFs at the scales from 2 to 4 are set to emphasize similarity to enhance F-measure, while the parameters of the last CRF at the last scale 5 are set to emphasize proximity to further reduce MAE. Thus, the kernels of the CRFs at scale 2 to 4 are set as $\sigma_\alpha=60$, $\sigma_\beta=5$, $\sigma_\gamma=3$ to emphasize more on the similarity, while the kernels of the CRFs at scale 5 are defined as $\sigma_\alpha=1$, $\sigma_\beta=10$, $\sigma_\gamma=10$. $T$ is set as 3 for efficient computation. 

\par\noindent\textbf{Message-passing.} We train the deep unified CRF model by involving variable combinations of message-passing within the inference. As in Table~\ref{tbl:f_s_passing}, the baseline is the backbone CNN based on the enhance HED structure and we evaluate the F-measure of the output prediction map $\textbf{o}^5$. By adding a Dense-CRF for post processing, the F-measure is 0.902. Then, the message-passing comparisons are conducted by implementing pairwise terms in Eq.~\ref{eq:energy} to the cascade CRFs architecture for joint model training. 1) By adding message-passing between predictions, the F-measure rises to 0.921. Also, by joint training CRF through back propagation, the prediction map from the baseline framework improves from 0.884 to 0.899. 2) By adding message-passing between features, the F-measure is 0.904, with 2\% increase to the baseline output. Finally, by adding feature-feature, feature-prediction and prediction-prediction messages passing, the F-measure further improves to 0.928.


\begin{table}[t]
\small
\begin{center}
    \begin{tabular}{l|c}
    \toprule 
    Method	& F-measure\\\midrule
    Baseline &   0.884 \\
    Baseline + Dense-CRF (post-processing)~\cite{krahenbuhl2011efficient} & 0.902 \\
    Baseline + CRF (/w P) (backbone output)& 0.899 \\
    Baseline + CRF (/w P) (CRF output)& 0.921\\
    Baseline + CRF (/w F) (CRF output) & 0.904 \\
    Baseline + CRF (/w P \& F) (CRF output)& 0.928\\\bottomrule
    \end{tabular}
    \vspace{5pt}
    \caption{F-measure of the estimated prediction map $\textbf{o}^5$ by implementing pairwise terms in Eq.~\ref{eq:energy} to the deep unified CRF model for message-passing comparisons. ``P'' refers to message-passing between predictions, ``F'' means message-passing between features, and ``/w P \& F'' means CRF with feature-feature, feature-prediction and prediction-prediction messages passing.}
    \vspace{-10pt}
    \label{tbl:f_s_passing}
\end{center}
\end{table}

\begin{figure}[!t]
\begin{center}
\scalebox{1}{\includegraphics[width=3.3in]{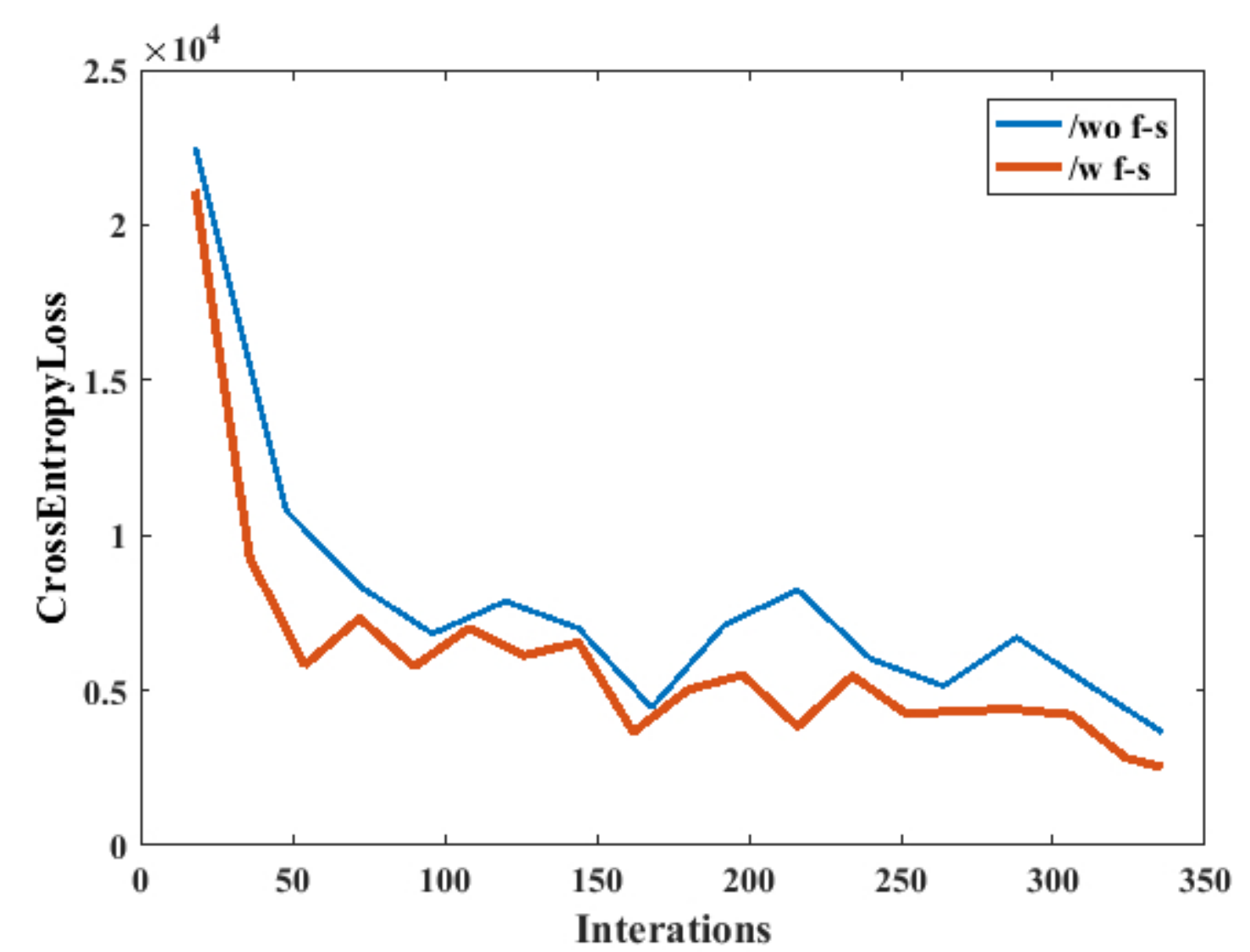}}
\caption{Training loss with (``/w f-s'') and without (``/wo f-s'') feature-prediction message-passing.}
\vspace{-18pt}
\label{fig:loss}
\end{center}
\end{figure}

\begin{figure*}[t]
\begin{center}
\scalebox{1}{\includegraphics[width=6.9in]{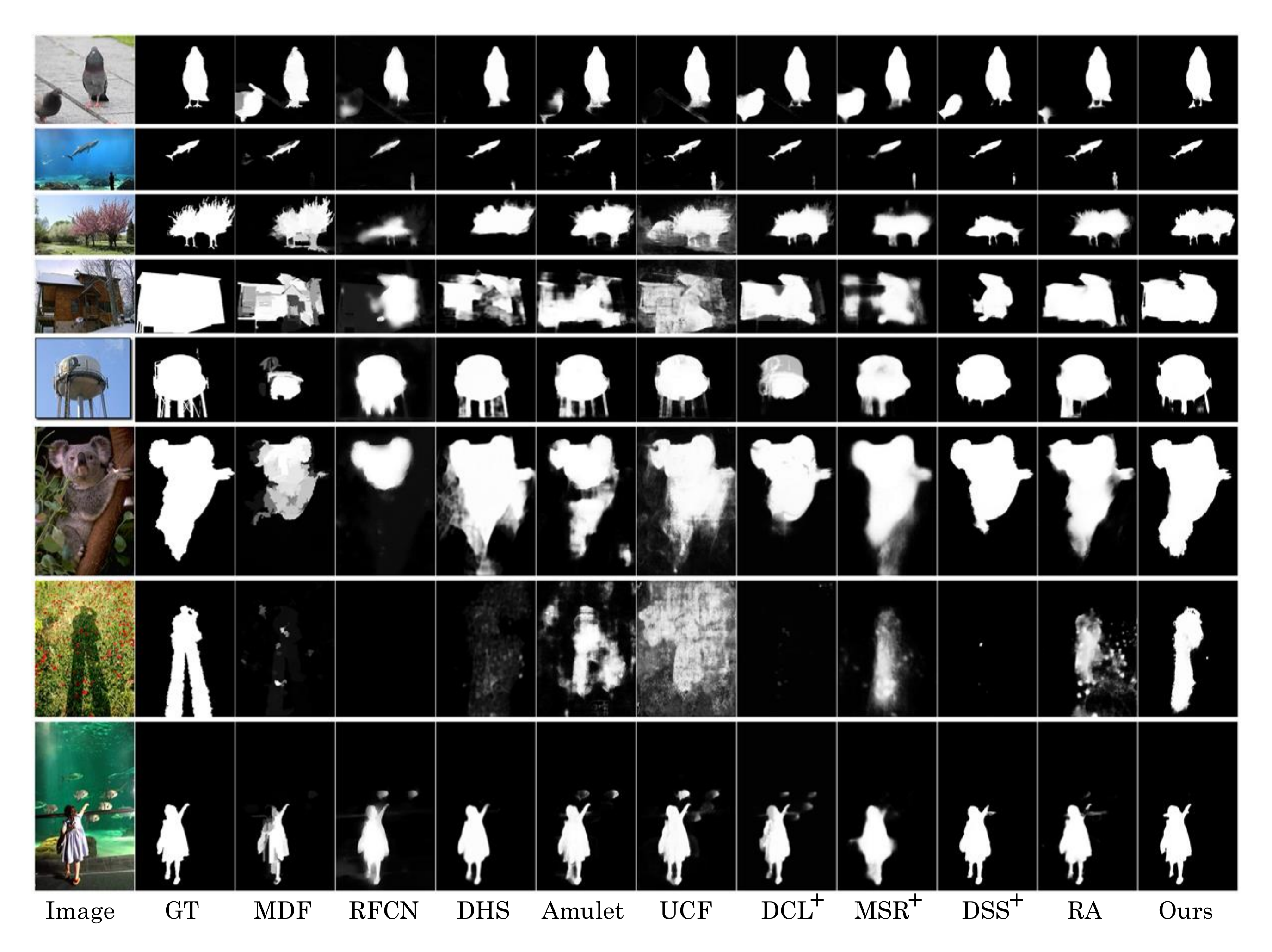}}
\vspace{-10pt}
\caption{Examples of saliency maps from MDF~\cite{li2015visual}, RFCN~\cite{wang2016saliency}, DHS~\cite{liu2016dhsnet}, Amulet~\cite{zhang2017amulet}, UCF~\cite{zhang2017learning}, DCL~\cite{li2016deep}, MSR~\cite{li2017instance}, DSS~\cite{hou2017saliency}, RA~\cite{chen2018reverse} and the proposed deep unified CRF model. ``$+$'' marks the models utilizing Dense-CRF~\cite{krahenbuhl2011efficient} for post-processing.}
\label{fig:example}
\end{center}
\end{figure*}

Further, Figure~\ref{fig:loss} plots the training loss by the cascade CRFs architecture with and without feature-prediction message-passing respectively. Clearly, building connections between features and predictions facilitates more efficient model training. The running time of the cascade CRFs architecture is similar to DSS model with Dense-CRF, with the same parameter settings for Gaussian kernels, taking approximately 0.48s per image.

\subsection{Cross Dataset Evaluation}
For comprehensive analysis, the proposed deep unified CRF model is compared with ten state-of-the-art saliency models including DRFI~\cite{jiang2013salient}, MDF~\cite{li2015visual}, RFCN~\cite{wang2016saliency}, DHS~\cite{liu2016dhsnet}, Amulet~\cite{zhang2017amulet}, UCF~\cite{zhang2017learning}, DCL~\cite{li2016deep}, MSR~\cite{li2017instance}, DSS~\cite{hou2017saliency}, RA~\cite{chen2018reverse}. All the models are CNN-based approaches except the DRFI model. All the implementations are based on public codes and suggested settings by the corresponding authors. Table~\ref{tbl:f_m} lists the max F-measure and MAE of the ten saliency models and the proposed deep unified CRF model over six datasets. It is observed that the deep unified CRF model results in better F-measure and significantly reduced MAE. Compared to DCL$^+$~\cite{li2016deep}, MSR~\cite{li2017instance} and DSS$^+$~\cite{hou2017saliency} that apply Dense-CRF~\cite{krahenbuhl2011efficient} as a post-processing method, the proposed jointly trained cascade CRFs effectively improve the performance. Figure~\ref{fig:example} presents saliency maps from the compared models and the proposed deep unified CRF model.



\subsection{Noise} 
Besides horizontal flipping for data augmentation, we explore other noise adding methods for robust model training. We add one CRF block with only prediction-prediction message-passing at scale 5 and jointly train with CNN for 10 epochs, and test the F-measure of the prediction map ${\textbf{o}}^5$ with various noise adding methods, on ECSSD dataset. Data augmentation with horizontal flipping results in F-measure as 0.910.

Firstly, we enlarge the training sets with both the horizontal flipping and the vertical flipping, the F-measure slightly decreases to 0.905. This may because that the symmetry properties for salient objects mostly apply to horizontal directions. Also, we add noises to the images, \ie, blurring, sharpening, Gaussian noise, and inversion, for data augmentation. The training takes much longer time and the F-measure is 0.896. Different from the detection and recognition tasks, the salient object detection tasks attach more importance to the smoothness of the resulted saliency maps. This may be the reason why additional augmentation does not result in a better model.

\section{Conclusion} \label{sec:conclusion}
This paper presents a novel deep unified CRF model. Firstly, we jointly formulate the continuous features and the discrete predictions into a unified CRF, which provides explainable solutions for the features, the predictions, and the interactions among them. Secondly, we infer mean-field approximations for highly efficient message-passing, leading to distinct model formulation, inference, and neural network implementation. Experiments show that the proposed cascade CRFs architecture results in highly competitive performance and facilitates more efficient model training.

{\small
\bibliographystyle{ieee_fullname}
\bibliography{egbib}
}

\end{document}